\pdfoutput=1

\documentclass[11pt]{article}

\usepackage[final]{acl}

\usepackage{times}
\usepackage{latexsym}

\usepackage[T1]{fontenc}

\usepackage[utf8]{inputenc}
\usepackage{amsmath}
\usepackage{microtype}
\usepackage{xspace}

\newcommand{\ourmethod}{CBS\xspace}

\newcommand{\baseline}{Default\xspace}

\usepackage{inconsolata}

\usepackage{graphicx}

\usepackage{multirow}
\usepackage{subfigure}
\title{Stay Tuned: An Empirical Study of the Impact of Hyperparameters on LLM Tuning in Real-World Applications}

\author {
Alon Halfon\thanks{\ \ These authors equally contributed to this work.}, Shai Gretz\footnotemark[1], Ofir Arviv\footnotemark[1], {\bf Artem Spector}, {\bf Orith Toledo-Ronen}, {\bf Yoav Katz}, \\ {\bf Liat Ein-Dor},{\bf Michal Shmueli-Scheuer}, {\bf Noam Slonim}\\\\
 IBM Research \\
 \
 \{alonhal,avishaig,artems,oritht,katz,liate,shmueli,noams\}@il.ibm.com,\\
 ofir.arviv@ibm.com
}

\begin{document}
\maketitle
\begin{abstract}
Fine-tuning Large Language Models (LLMs) is an effective method to enhance their performance on downstream tasks. However, choosing the appropriate setting of tuning hyperparameters (HPs) is a labor-intensive and computationally expensive process. Here, we provide recommended HP configurations for practical use-cases that represent a better starting point for practitioners, when considering two SOTA LLMs and two commonly used tuning methods. We describe Coverage-based Search (CBS), a process for ranking HP configurations based on an offline extensive grid search, such that the top ranked configurations collectively provide a practical robust recommendation for a wide range of datasets and domains. We focus our experiments on Llama-3-8B and Mistral-7B, as well as full fine-tuning and LoRa, conducting a total of $> 10,000$ tuning experiments. Our results suggest that, in general, Llama-3-8B and LoRA should be preferred, when possible. Moreover, we show that for both models and tuning methods, exploring only a few HP configurations, as recommended by our analysis, can provide excellent results in practice, making this work a valuable resource for practitioners.
\end{abstract}

\section{Introduction}

Fine-tuning Large Language Models (LLMs) is an effective method to enhance their performance by adapting them to specific domains and tasks \cite{shi2023specialistgeneralistinstructiontuning}. This approach is particularly valuable in real-world enterprise scenarios, where there is often a need to address specific downstream tasks using available data, such as the company's proprietary data.

Recently released base-models, such as Gemma \cite{gemmateam2024gemmaopenmodelsbased}, Llama \cite{touvron2023llamaopenefficientfoundation,touvron2023llama2openfoundation}, and Mistral \cite{jiang2023mistral7b}, claim ease of 
fine-tuning across various tasks \cite{zhao2024loraland310finetuned}. However, comprehensive studies of these models in the context of fine-tuning are still limited, leaving several 
important 
questions less explored. In this paper, we focus on the role of hyperparameter optimization (HPO) in the fine-tuning process of LLMs, 
and provide detailed and concrete recommendations for HP values, aiming to save practitioners time and computational resources. 
We present 
Coverage-based Search (\ourmethod), which leverages an extensive grid search for highlighting
an effective HP recommendation, as well as the ability to expand to 
a few promising HP configurations that collectively suggest high performance across diverse datasets and tasks. 

For the purpose of providing these recommendations we conduct a comprehensive systematic study, focusing on practical scenarios where relatively small 
training data 
are available for tuning. We examine prominent tasks such as classification, summarization, and contextual question-answering (CQA) across various domains. Our study considers two leading LLMs, Llama-3-8B \cite{llama3modelcard} and Mistral-7B-v0.3 \cite{jiang2023mistral7b}, as well as two commonly used fine-tuning methods: full fine-tuning (FFT) and LoRA \cite{hu2021lora}.

Our main contributions are as follows:
\begin{enumerate}
    \item Recommended HP configurations for tuning, optimized per model and tuning method.
    \item Analysis of the differences between Llama-3-8B and Mistral-7B-v0.3, as well as between LoRA and FFT, across 3 real-world tasks in practical scenarios.
    \item Analysis of the potential gain, accumulated by considering additional HP configurations suggested by our analysis. 
\end{enumerate}

\section{Related Work}
HPO is an established research area, 
known for its critical role in enhancing model performance \cite{yu2020hyperparameteroptimizationreviewalgorithms, jin2022hyperparameterimportancemachinelearning}. 
The most straightforward approach for HPO 
is a grid search over the exponential 
space of HP values 
\cite{bergstra2012random}. Since 
grid search is computationally demanding, 
a large volume of research has been focused on developing and evaluating 
more efficient HPO methods 
\cite{Bergstra2011AlgorithmsFH,Swersky2013MultiTaskBO,snoek2012practicalbayesianoptimizationmachine,liu-wang-2021-empirical}, 
while others consider 
over which HPs one should focus on 
\cite{gkouti2024itrymultipleoptimizers,zhang-duh-2020-reproducible,huang2024hyperparameterlosslandscapesmachine}.
A few recent studies, described next, aimed to provide concrete recommendations for HP settings. However, these works typically considered a limited collection of datasets, tasks, or HPs.
\citet{j2024finetuningllmenterprise} who fine-tuned the Llama-2 model on RAG and Code generation tasks, compared FFT 
and LoRA, 
and provided some general recommendations; however, their evaluation was limited to a single dataset per task, considering a small test set, and exploring a limited set of HP configurations. 
\citet{zhang2024scalingmeetsllmfinetuning} examined the effects of scaling model size, data size, and PEFT parameters on machine translation and multilingual summarization, and found that larger data size improved performance, while scaling PEFT parameters was ineffective. 
\citet{tribes2024hyperparameteroptimizationlargelanguage} evaluated Llama-2 7B using LoRA tuning, exploring configurations of rank, alpha, dropout, and learning rate on instruction datasets. Utilizing black box optimization techniques, they identified that a learning rate around $10^{-3.5}$ yielded the best results, while other HPs showed no decisive optimal values.
In contrast, the present work considers both FFT and LoRA, for two SOTA models, using a comprehensive grid search across a large number of HP configurations, for a wide range of datasets, covering multiple domains and tasks. 
Thus, we expect the recommendations suggested here 
to provide a significant added value on top of previous research in this area.

\section{Experimental Setup} 

\begin{figure}[ht]
\begin{center}
\includegraphics[bb=0bp 50bp 642bp 378bp,clip,width=1.0\columnwidth]{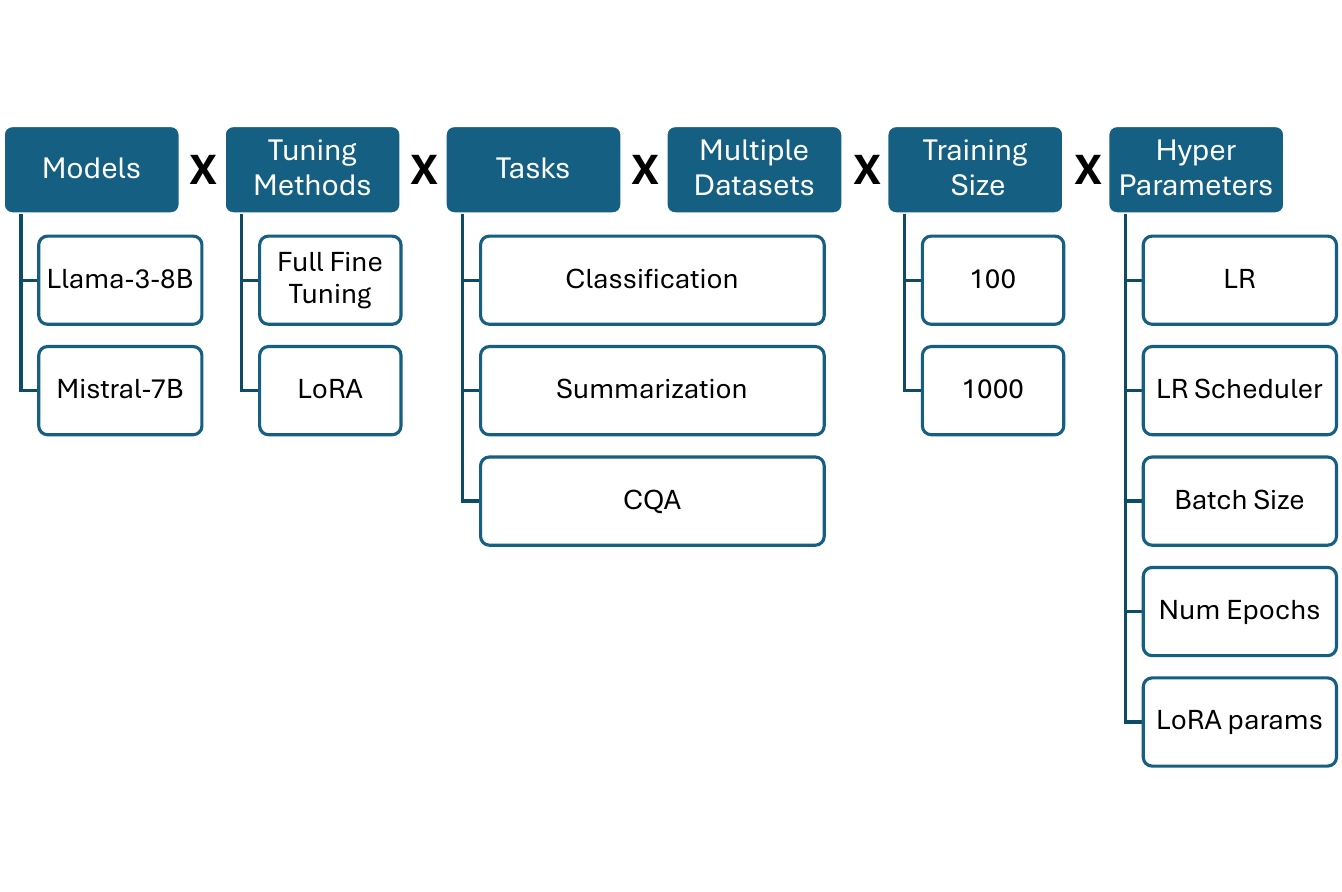}
\caption{Cartesian product defining our experimental setup. The evaluation is performed across two models, two tuning methods, three tasks with multiple datasets, two training set sizes, and over multiple HPs.}
\label{fig:evaluation_score}
\end{center}
\end{figure}

%
Our experimental setup is concisely depicted in Figure~\ref{fig:evaluation_score}. 
For each pair of model and tuning method 
we consider $3$ tasks, multiple datasets, and $2$ training sizes, 
as well as several 
HPs. For each of these HPs we consider 
multiple 
values, 
and apply a grid search over 
all the resulting 
HP configurations to identify the best one. 
Next, we 
dive deeper into each 
part of this setup. 

\subsection{Tasks and Datasets}
We 
consider $3$ 
tasks: text classification, text summarization, and contextual question answering (CQA). For text classification, we use $5$ 
multi-class datasets from various domains, with class counts ranging from $6$ to $100$. For text summarization, we utilize $5$ 
datasets from different domains, featuring diverse input and output lengths. 
For CQA, we include $3$ 
datasets, one of which (DoQA) consists of $3$ sub-datasets, one per domain, 
resulting in a total of $5$ 
datasets for this task.

We adhere to the original train/validation/test splits when available. Otherwise, we create train and validation splits, 
allocating a portion (which differs from dataset to dataset) of the training data to validation. 
%
Full details of the datasets are provided in Appendix~\ref{app_datasets}.




\subsection{Models}

There is a plethora of models that can be tuned over labeled data, and naturally not all can be covered 
with limited resources. 
Thus, we considered representatives of two of the most popular families of open-source models, restricting their size to 8B or less for practical reasons: Llama-3-8B and Mistral-7B-v0.3.

\subsection{Fine-Tuning Methods}

We explore 
two tuning techniques commonly used by practitioners: 
FFT and LoRA. FFT updates all model parameters, offering potential higher performance 
gain 
at a greater computational cost. LoRA reduces the number of learnable parameters by approximating weight matrices, significantly lowering computational overhead. Our study aims to identify recommended HP 
configurations for each approach. In addition, we compare the two methods to evaluate their effectiveness in terms of performance gains across different models and tasks.

\subsection{HP Search Space}
\label{sec:hp_search}
For FFT, we tune $4$ 
key HPs: 
learning rate (\emph{LR}), learning rate scheduler (\emph{LR Scheduler}), effective batch size\footnote{The effective batch size is a product of actual batch size, number of GPUs, and gradient accumulation steps.} (\emph{Batch}), and number of epochs (\emph{Epochs}).
For LoRA fine-tuning, we additionally explore LoRA-specific HPs: 
the rank (\emph{LoRA$_R$}) and scaling factor (\emph{LoRA$_\alpha$}), and fixing the scheduler.

We define a separate search space for each model and tuning method based on preliminary experiments, assuming that optimal settings vary across different models and tuning techniques. The full search space is detailed in the upper part of Table~\ref{tab:search_space}. This comprehensive search involves 
$96-288$ 
configurations 
for each tuple of model, tuning method, and dataset. 
Thus, in total, the results reported in this study are based on 
$> 10,000$ fine-tuning experiments.\footnote{Including preliminary experimentation and reproducing results.}

\subsection{Coverage-based Search (\ourmethod)}
\label{sec:our_method}

Next, we outline our 
approach to obtain recommended HP configurations using grid search results across multiple datasets. Our strategy leverages the diversity of tasks and domains, aiming for good 
coverage for unseen datasets. For each model and tuning method, we evaluate all $D$ datasets across the $3$ tasks, considering results obtained on training sizes $M=m1, m2$, using a set of possible HP configurations $C$. The goal is to identify a ranking of HP configurations such that the top 
configurations yield 
consistently good results for most datasets and training sizes.

First, for a given model, tuning method, dataset $d$, and training size $m$, we 
denote by 
$s(c)$ the score obtained by HP configuration 
$c$. The score is normalized w.r.t to the maximum score on $d$ using the same model, tuning method, and training size:
\[
s_n(c) = \frac{s(c)}{\max_{c \in C} s(c)}~. 
\]

For a given dataset $d$ and training size $m$, we denote the top configurations, $TC$, as the
configurations that receive a score that is at most 
$3$ percent lower than the best configuration:\footnote{We chose $3\%$ since it provided a good balance between having in $TC$ too common and too unique configurations.} 
\[
TC(d, m) = \{c \in C \mid s_n(c) > 0.97\}~. 
\]

We define $TC^*$ 
as the union of $TC(d, m)$ for all $(d,m)$ in 
$D \times M$. 
The score of each configuration $c$ in $TC^*$ is defined as follows - 
\[
S_n(c) = \sum_{d,m \in D \times M, c \in TC(d,m)} s_n(c)~. 
\]
This score essentially counts the number of times $c$ was selected in the top configurations $TC^*$.\footnote{We sum over $s_n(c)$ and not over the sign of $s_n(c)$ to break ties.}
Finally, we 
sort $TC^*$ according to the value of $S_n(c)$ in descending order.

HP configurations that work well for some datasets may not be optimal for other datasets. To take that into account, while still providing the practitioner with a small set of recommended HP configurations, we take the following approach.
First, we define $RankedAbove (c)$ as the set of configurations $c1 \in C$ s.t. $S_n(c1) > S_n(c)$. Next, when iterating over $TC^*$ we calculate for each $c$:
\[
coverage(c) =
\]
\[
\{ (d,m) \in D \times M \mid c \in TC(d, m),
\]
\[\underset{\forall c1 \in RankedAbove(c)}{c1 \notin TC(d, m)} \}~.
\]
That is, the set of $(d,m)$ for which $c$ provides a good result (i.e., $c \in TC(d, m)$) while higher-ranked configurations do not. 
We finally sort $TC^*$ by 
the size of 
$coverage(c)$ in descending order, to obtain a ranking over the HP configurations.

\begin{table*}[!ht]
    \centering
    \small
    \resizebox{\textwidth}{!}{%
    \begin{tabular}{|l|l|l|l|l|l|l|l|l|}
    \hline
        \textbf{Configuration} & \textbf{Model} & \textbf{Method} & \textbf{Batch} & \textbf{LR} & \textbf{Epochs} & \textbf{LR Scheduler} & \textbf{LoRA$_R$} & \textbf{LoRA$_\alpha$} \\ \hline\hline
        CBS & Mistral-7B-v0.3 & Full FT & [8, 32] & [5e-07, 1e-06, 5e-06, 1e-05] & [5, 10] & ['constant', 'cosine', 'linear'] & -- & -- \\ \hline
        CBS & Mistral-7B-v0.3 & LoRA & [8, 32] & [5e-05, 0.0001, 0.0005, 0.001] & [5, 10] & ['cosine']~ & [4, 32, 128] & [8, 64, 128] \\ \hline
        CBS & Llama-3-8B & Full FT & [8, 32] & [1e-06, 5e-06, 1e-05] & [5, 10] & ['constant', 'cosine', 'linear'] & -- & -- \\ \hline
        CBS & Llama-3-8B & LoRA & [8, 32] & [5e-05, 0.0001, 0.0005, 0.001] & [5, 10] & ['cosine'] & [4, 32, 128] & [8, 64, 128] \\ \hline \hline  
        \baseline & Mistral-7B-v0.3 & Full FT & [8] & [5e-05] & [10] & ['linear'] & -- & -- \\ \hline
        \baseline & Mistral-7B-v0.3 & LoRA & [1] & [1e-04] & [10] & ['linear'] & [128] & [64] \\ \hline
        \baseline & Llama-3-8B & Full FT & [8] & [5e-05] & [10] & ['linear'] & -- & -- \\ \hline
        \baseline & Llama-3-8B & LoRA & [4] & [1e-4] & [10] & ['linear'] & [32] & [8] \\ \hline
    \end{tabular}
    }
    \caption{CBS search space and the \baseline HP configurations. Where an HP has a single value no search was done.}
    \label{tab:search_space}
\end{table*}

\section{Evaluation Details}
\subsection{Data}
We focus on practical scenarios where training data is typically limited. 
Thus, for each dataset we evaluate two variants of training data sizes: $100$ and $1000$, sampled at random. The validation and test set sizes are $1000$ for classification, $500$ for summarization, and $329-500$ 
(depending on availability) 
for CQA. Note, 
both training data sizes are used to identify the recommended HP configurations, and we do not optimize these recommendations for specific training sizes.
For downloading and processing the datasets we use the Unitxt library~\cite{bandel-etal-2024-unitxt}: a collaborative framework for unified textual data processing and evaluation which allows easy formatting, sharing, and reproducibility of LLM evaluation results. For classification, we ensure each class has at least one train sample to prevent missing classes in the train set. \footnote{The Unitxt recipes used in our experiments, which allow for full reproduction of our data, will be released upon publication.}

\subsection{Training}
We use the SFTTrainer from HuggingFace Transformers library \cite{wolf2020huggingfacestransformersstateoftheartnatural} with PyTorch FSDP. Each tuning and inference process utilizes either a single NVIDIA A100 with 80GB or a pair of them, operating at FP16 precision.

\subsection{Methods to Select HP Configurations}
\label{sec:evaluation}
For each model and tuning method, we aim to provide HP recommendations for the practitioner. To determine the quality of these 
recommendations, we consider the following recommendation alternatives.

\textbf{\baseline}. A 
common practice is to retrieve HP recommendations from publicly available sources. For LoRA, we evaluate official HP recommendations for tuning Llama\footnote{\url{https://github.com/meta-llama/llama-recipes/tree/main}} and Mistral.\footnote{\url{https://github.com/mistralai/mistral-finetune/blob/main/example/7B.yaml}} For FFT, we could not find HP recommendations in our 
literature and online search. Thus, we use the default parameters provided by HuggingFace.\footnote{\url{https://huggingface.co/docs/trl/en/sft_trainer}} 
The HPs defined by each \baseline configuration can be found at the bottom of Table \ref{tab:search_space}.\footnote{The \baseline recommendations suggested using $3$ epochs, we used $10$ as we wanted to strengthen their results.}

\textbf{\ourmethod Leave-one-dataset-out (LOO)}. We evaluate the approach presented in Section~\ref{sec:our_method} in a LOO fashion, to simulate the benefit of its recommended HP configuration on new datasets. Note that we consider only the 
single top 
configuration entailed by the CBS ranking. We denote this method as \textbf{CBS\_1}.
For each held-out dataset, $d_h$, we calculate the top HP configuration obtained by running \ourmethod on $D \backslash d_h$. We then take the score achieved by using this configuration on the test set of $d_h$.

\textbf{Upper Bound}. We optimize the HPs of each dataset separately on its validation set, and report the score on the test set. In other words, this is a full grid search for each dataset, which is quite demanding and typically not feasible in practical scenarios. 

We report 
micro-f1 for text classification datasets, and rougeL for summarization and CQA datasets. For reporting the performance on each task we report macro-average over the respective datasets. \footnote{We will share with the community the 
 complete results 
 of our grid search on all datasets upon publication.} 

\section{Results and Analysis}

\subsection{HP Recommendations}

Table \ref{tab:hp_configurations_comparisons_full} presents the average performance of \ourmethod compared to the \baseline configuration and the upper bound for FFT. The results show that CBS\_1 outperforms the \baseline method by a large margin across all tasks, models, and train sizes. 

The results of the same experiment with LoRA are shown in Table \ref{tab:hp_configurations_comparisons_LORA}. Here, for LLama-3-8B, the performance of CBS\_1 is either comparable or slightly better than the \baseline configurations, indicating that the recommendations published for Llama-3-8B are beneficial.
In contrast, for Mistral-8B-v0.3, CBS outperforms the baseline recommendation by a large margin in all cases. \textbf{Thus, for both models, the 
CBS\_1 
configuration can be considered a new HP recommendation for both FFT and LoRA in the considered region of small training size.}

\begin{table*}[!ht]
    \centering
    \resizebox{\textwidth}{!}{%
    \begin{tabular}{|l|c|c|c|c|c|c|c|c|c|c|c|c|}
    \hline
        & \multicolumn{6}{c|}{\textbf{Llama-3-8B}} & \multicolumn{6}{c|}{\textbf{Mistral-7B-v0.3}} \\ \hline
        \textbf{Task} & \multicolumn{3}{c|}{\textbf{100}} & \multicolumn{3}{c|}{\textbf{1000}} & \multicolumn{3}{c|}{\textbf{100}} & \multicolumn{3}{c|}{\textbf{1000}}\\ \cline{2-13}
        ~ & \baseline & CBS\_1 & Upper Bound & \baseline & CBS\_1 & Upper Bound & \baseline & CBS\_1 & Upper Bound & \baseline & CBS\_1 & Upper Bound \\ \hline
        Classification & 47.96 & 70.17 & 72.25 & 72.11 & 80.52 & 81.48 & 3.94 & 58.56 & 68.58 & 14.73 & 65.89 & 78.14 \\ \hline
        Summarization & 19.45 & 27.50 & 28.44 & 23.06 & 27.32 & 29.28 & 12.72 & 27.28 & 28.09 & 16.86 & 27.17 & 29.07 \\ \hline
        CQA & 40.97 & 54.91 & 56.71 & 53.97 & 66.57 & 67.64 & 25.44 & 53.17 & 54.85 & 35.59 & 64.45 & 66.69 \\ \hline
    \end{tabular}
    }
    \caption{Comparing HP configurations in \textbf{FFT}.}
    \label{tab:hp_configurations_comparisons_full}
\end{table*}

\begin{table*}[!ht]
    \centering
    \resizebox{\textwidth}{!}{%
    \begin{tabular}{|l|c|c|c|c|c|c|c|c|c|c|c|c|}
    \hline
        & \multicolumn{6}{c|}{\textbf{Llama-3-8B}} & \multicolumn{6}{c|}{\textbf{Mistral-7B-v0.3}} \\ \hline
        \textbf{Task} & \multicolumn{3}{c|}{\textbf{100}} & \multicolumn{3}{c|}{\textbf{1000}} & \multicolumn{3}{c|}{\textbf{100}} & \multicolumn{3}{c|}{\textbf{1000}}\\ \cline{2-13}
        ~ & \baseline & CBS\_1 & Upper Bound & \baseline & CBS\_1 & Upper Bound & \baseline & CBS\_1 & Upper Bound & \baseline & CBS\_1 & Upper Bound \\ \hline
        Classification & 68.92 & 69.38 & 74.43 & 79.70 & 79.85 & 81.84 & 49.35 & 63.76 & 70.63 & 54.58 & 71.46 & 79.54 \\ \hline
        Summarization & 25.80 & 27.25 & 28.85 & 26.79 & 26.96 & 29.36 & 23.25 & 26.29 & 28.09 & 25.09 & 27.18 & 29.16 \\ \hline
        CQA & 54.58 & 54.26 & 57.41 & 64.12 & 66.10 & 68.49 & 47.26 & 53.91 & 56.89 & 55.19 & 61.29 & 68.02 \\ \hline
    \end{tabular}
    }
    \caption{Comparing HP configurations in \textbf{LoRA}.}
    \label{tab:hp_configurations_comparisons_LORA}
\end{table*}

\subsection{Upper Bound vs. CBS\_1}
\label{sec:benefits_of_full}
In Tables~\ref{tab:hp_configurations_comparisons_full} and~\ref{tab:hp_configurations_comparisons_LORA} we consider the gap between the recommendation provided by our CBS\_1 approach, in LOO mode, compared to selecting the best configuration found via a comprehensive HP search over the validation set of the individual dataset.
Evidently, for Llama-3-8B our approach is quite close to the upper bound, while for Mistral-7B-v0.3 the gap is more evident. 
However, we note that in practice, full HP search over the validation set is often not feasible. 


\subsection{Tuning Methods}

In general, FFT is known to demand more computational resources compared to LoRA. This is particularly true in our experimental setting with relatively smaller train data size. Thus, a pertinent question arises: what is the performance gain from FFT, and is it worth the increased computational cost? An examination of the results in Tables~\ref{tab:hp_configurations_comparisons_full} and \ref{tab:hp_configurations_comparisons_LORA} reveals that, across all configurations of \baseline, \ourmethod, and the upper bound, in most cases, there is no significant performance gain with FFT. This observation aligns with findings previously reported in~\citet{zhang2024scalingmeetsllmfinetuning} under different settings. Based on this analysis, 
\textbf{when using small training data, our recommendation is 
to use LoRA, as it requires lower hardware resources while delivering similar or even superior performance compared to FFT, for both models and across all tasks.}


\subsection{Models}
Overall, there is a clear advantage for LLama-3-8B over Mistral-7B-v0.3 across all dimensions. This finding is in line with the ranking in the Fine-tuning Leaderboard where Llama-3-8B and Mistral-7B-v0.3 are ranked first and fifth, respectively \cite{zhao2024loraland310finetuned}.\footnote{\url{https://predibase.com/fine-tuning-index}}

\subsection{Train Data Size}
As expected, moving from $100$ to $1000$ train samples improves the results across all tasks, models, and tuning methods. Notice, that despite the differences in performance scores, 
the overall trends between the considered configurations (\baseline, \ourmethod, upper bound) are qualitatively similar across the data sizes.

\subsection{Impact of Exploring Multiple CBS Recommendations}
Next, we examine the impact of exploring 
more than one configuration 
from our recommended ranked list. 
To that 
end, we expand the evaluation of \ourmethod described in Section~\ref{sec:evaluation} to consider additional HP recommendations beyond the top one. 
From a practical perspective. the budget is therefore defined 
as the number of configurations in $TC^*$ we evaluate.

For each $d_h$ (the held-out dataset) and training size $m$, we iterate over $c \in TC^*$ according to the ranking induced by $coverage(c)$. Assuming we have a budget of size $k$, we consider the top $k$ configurations in $TC^*$, and evaluate $s(c)$ of each configuration on the validation set of $h_d$. We mark the configuration with the highest $s(c)$ as $c_{best}$. Finally, we calculate $s_n(c_{best})$ (the normalized score, see Section~\ref{sec:our_method}) on the test set of $d_h$. We then average these scores over all held-out datasets and training sizes.


\begin{figure}[ht]
\begin{center}
\includegraphics[width=0.97\columnwidth]{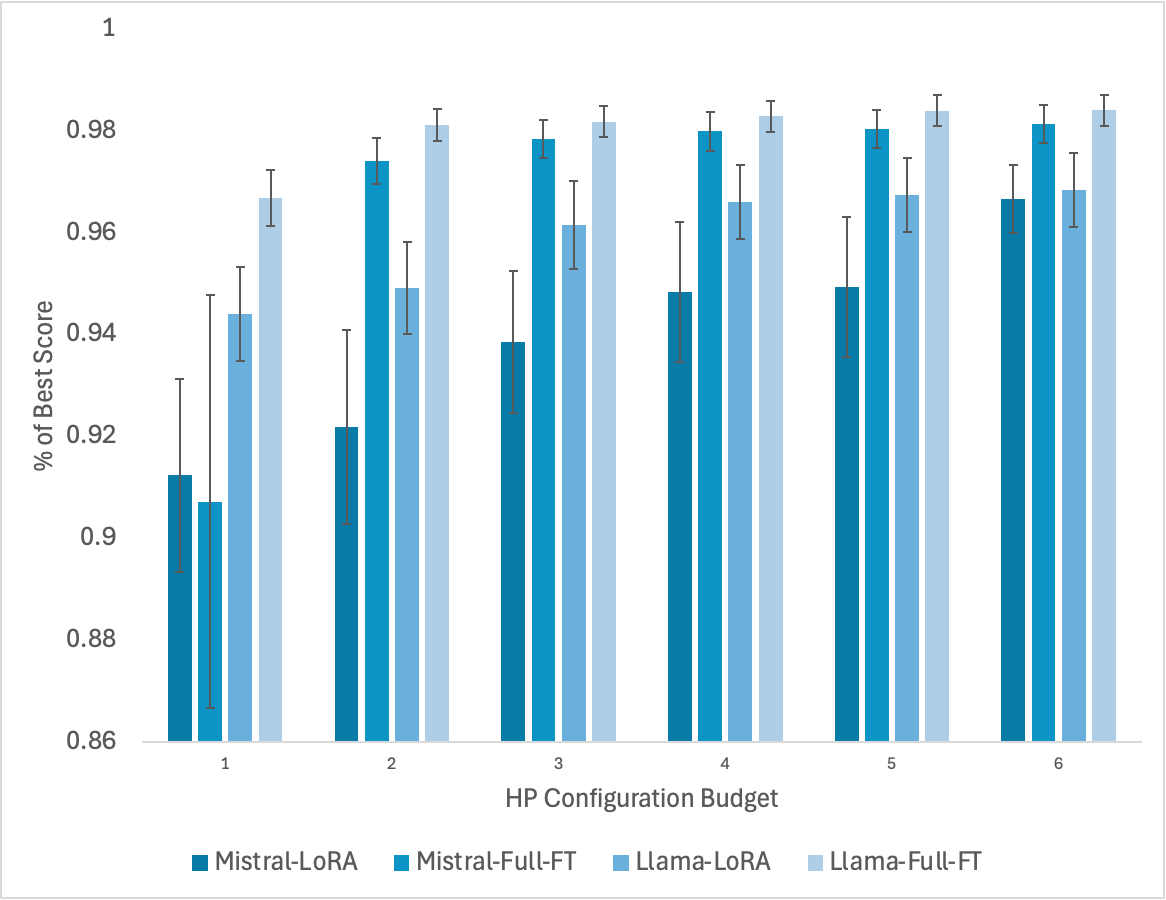}
\caption{Effect of increasing HP configuration budget. Y-axis denotes macro-averaged scores over all datasets and training sizes normalized w.r.t the upper bound score obtained by the respective model and tuning method (i.e., $s_n(c))$.}
\label{fig:hpo_budget}
\end{center}
\end{figure}

We present in Figure~\ref{fig:hpo_budget} the average performance as defined above as a function of 
the configuration budget.
Evidently, 
all methods benefit from adding at least 
one additional HP configuration. However, for all setups besides Mistral-LoRA, there is almost no change in performance beyond the top $4$ configurations. The highest benefit is achieved for FFT with Mistral-7B-v0.3, where there is a large increase when moving from budget $1$ to $2$. Still, LoRA remains the superior tuning method even with an increased budget.

\subsection{Recommendation for the Practitioner}

\begin{table*}[ht]
\small
\centering
\begin{tabular}{|l|l|c|c|c|c|c|c|c|}
\hline
\textbf{Model} & \textbf{Method} & \textbf{Rank} & \textbf{LR} & \textbf{LR Scheduler} & \textbf{Batch} & \textbf{Epochs} & \textbf{LoRA$_R$} & \textbf{LoRA$_\alpha$} \\ \hline\hline
Mistral-7B-v0.3 & Full-FT & 1 & 5e-06 & linear       & 8 & 5 & -- & -- \\
& & 2 & 1e-06 & constant		& 8	& 5 & -- & -- \\
& & 3 & 5e-06 & cosine		& 8	& 5 & -- & -- \\
& & 4 & 5e-06 & cosine		& 32 & 5 & -- & -- \\ \hline\hline
Mistral-7B-v0.3 & LoRA &  1 &  5e-05 &	cosine	& 32 &	5 & 128	& 128 \\
& & 2 & 5e-05 &	cosine	& 8	& 5	& 32 & 128 \\
& & 3 & 5e-05 &	cosine	& 32 & 	5 &	128 & 8 \\
& & 4 & 5e-05 &	cosine	& 8	& 5	& 4	& 64 \\ \hline\hline
Llama-3-8B & Full-FT & 1 & 1e-05 & cosine	& 8	& 5 & -- & -- \\
& & 2 & 5e-06 &	cosine	& 8	& 5 & -- & -- \\
& & 3 & 1e-05 &	constant	& 8	& 10 & -- & -- \\
& & 4 & 1e-05 &	linear	& 8	& 5 & -- & -- \\ \hline\hline
Llama-3-8B & LoRA  & 1 & 5e-05	& cosine	& 8	& 5	& 32	& 128 \\
& & 2 & 1e-04	& cosine	& 32	& 5	& 128	& 64 \\
& & 3 & 1e-04	& cosine	& 32	& 5	& 4	& 8 \\
& & 4 & 1e-04	& cosine	& 8	& 5	& 128	& 64 \\ \hline\hline
\end{tabular}
\caption{CBS HP recommendations.}
\label{tab:recommendations_table}
\end{table*}
Based on our experiments, we created practical HP recommendations for each model and tuning method. Table~\ref{tab:recommendations_table} shows the $4$ top-ranked configurations, and we suggest to use these HP configurations 
in order, 
according to the available budget. 
As shown in Figure~\ref{fig:hpo_budget}, using these $4$ configurations, or even less, is expected to yield results nearly equivalent to full grid search over the HP space. 

\section{Conclusions}

To effectively fine-tune LLMs, it is essential to use a proper HP configuration, aligned with the model and tuning method at hand. Our work aims to contribute to the understanding of this aspect by providing practitioners with recommended HP configurations for two leading models and two tuning methods. These recommendations represent the outcome of the analysis of the results of more than $10,000$ fine-tuning experiments, across a large collection of datasets, representing different tasks and domains. Furthermore, we provide comparative analysis between Llama-3-8B and Mistral-7B-0.3, and between LoRA and FFT, indicating in both cases the advantage of the former option. Taken together, we believe our results should be of significant practical value for practitioners in the field. 

In future work, we plan to expand our analysis by considering additional HPs such as warmup ratio and weight decay. We will also compare our CBS approach with more advanced HPO algorithms. Finally, we intend to periodically update this work with new recommendations for HP configurations for additional models and tuning methods, aiming to further establish this work as a valuable resource in practice.


\bibliography{acl_latex}

\appendix

\section{Datasets}
\begin{table*}[hbt!]
    \small
    \centering
    \begin{tabular}{|p{6cm}|p{4.5cm}|l|}
    \hline
    \textbf{Dataset} & \textbf{Description} & \textbf{Task}  \\
    \hline \hline
    Head-QA \cite{vilares-gomez-rodriguez-2019-head} & Healthcare questions & Classification  \\
    \hline
    20 Newsgroups \cite{lang95newsweeder} & News discussions & Classification  \\
    \hline
    TREC \cite{li-roth-2002-learning}& Questions & Classification  \\
    \hline
    Banking77 \cite{casanueva-etal-2020-efficient}& Queries to banking chatbot & Classification  \\
    \hline
    LEDGAR \cite{chalkidis-etal-2022-lexglue}& Legal clauses & Classification  \\
    \hline \hline
    TL;DR \cite{tldr} & Reddit posts & Summarization  \\
    \hline
    CNN-DM \cite{see-etal-2017-get}  & News articles & Summarization  \\
    \hline
    Xsum \cite{Narayan2018DontGM} & News articles & Summarization \\
    \hline
    XL-Sum \cite{hasan-etal-2021-xl} & News articles & Summarization  \\
    \hline
    BillSum \cite{kornilova-eidelman-2019-billsum}& Congress bills & Summarization  \\
    \hline \hline
    CLAP NQ \cite{rosenthal2024clapnq} & Wikipedia (Long-form answers) & CQA  \\
    \hline
    DoQA \cite{campos2020doqa} & Cooking, travel and movies  & CQA\\
    \hline
    Open Australian Legal QA \cite{butler-2023-open-australian-legal-dataset} &  Legal Corpus  & CQA\\
    \hline
    \end{tabular}
    \caption{Datasets used in our evaluation.}
    \label{datasets_table}
\end{table*}
\label{app_datasets}
The datasets considered in this work are presented in Table~\ref{datasets_table}.

\end{document}